\documentclass{article}
\pdfoutput=1 
\usepackage{spconf,amsmath,graphicx,bm,amssymb}
\usepackage{hyperref}
\usepackage{xcolor}
\usepackage[font=small,labelfont=bf]{caption}
\usepackage[square,sort&compress,comma,numbers]{natbib}
\usepackage{float}
\usepackage{dblfloatfix}
\usepackage[]{color-edits}

\title{End-to-end AI-based MRI reconstruction and lesion detection pipeline for evaluation of deep learning image reconstruction}

\name{Ruiyang Zhao$^{1,2,*}$, Yuxin Zhang$^{1,2,*}$, Burhaneddin Yaman$^{1,3,}$\sthanks{These authors have equal contribution to this work.}, Matthew P. Lungren$^{1,4}$, Michael S. Hansen$^1$}
\address{$^1$ Microsoft Research, Health Futures\\
$^2$ University of Wisconsin-Madison, Department of Medical Physics \\
$^3$ University of Minnesota, Department of Electrical and Computer Engineering \\
$^4$ Stanford University, Department of Radiology}

\begin{document}
%
\maketitle
\begin{abstract}
Deep learning techniques have emerged as a promising approach to highly accelerated MRI. However, recent reconstruction challenges have shown several drawbacks in current deep learning approaches, including the loss of fine image details even using models that perform well in terms of global quality metrics. In this study, we propose an end-to-end deep learning framework for image reconstruction and pathology detection, which enables a clinically aware evaluation of deep learning reconstruction quality. The solution is demonstrated for a use case in detecting meniscal tears on knee MRI studies, ultimately finding a loss of fine image details with common reconstruction methods expressed as a reduced ability to detect important pathology like meniscal tears. Despite the common practice of quantitative reconstruction methodology evaluation with metrics such as SSIM, impaired pathology detection as an automated pathology-based reconstruction evaluation approach suggests existing quantitative methods do not capture clinically important reconstruction outcomes.
\end{abstract}
\section{Introduction} 
\label{sec:intro}
Compared to other cross-sectional medical imaging modalities, MRI is traditioanlly slow and inefficient which has motivated the use of accelerated MRI techniques. Parallel imaging \cite{Grappa,Sense} and compressed sensing \cite{lustig} are some of the conventional approaches used in clinical studies, but their acceleration rate is limited due to noise amplification and artifacts. In recent years, deep learning (DL) techniques have emerged as an alternative technique for accelerated MRI due to their improved reconstruction quality compared to conventional approaches at high rates \cite{WangDLMRI,Schlemper,Hammernik,Hemant,yaman_SSDU_MRM,Knoll_SPM,LeslieYing_SPM}.  

\begin{figure}
\includegraphics[width=0.5\textwidth]{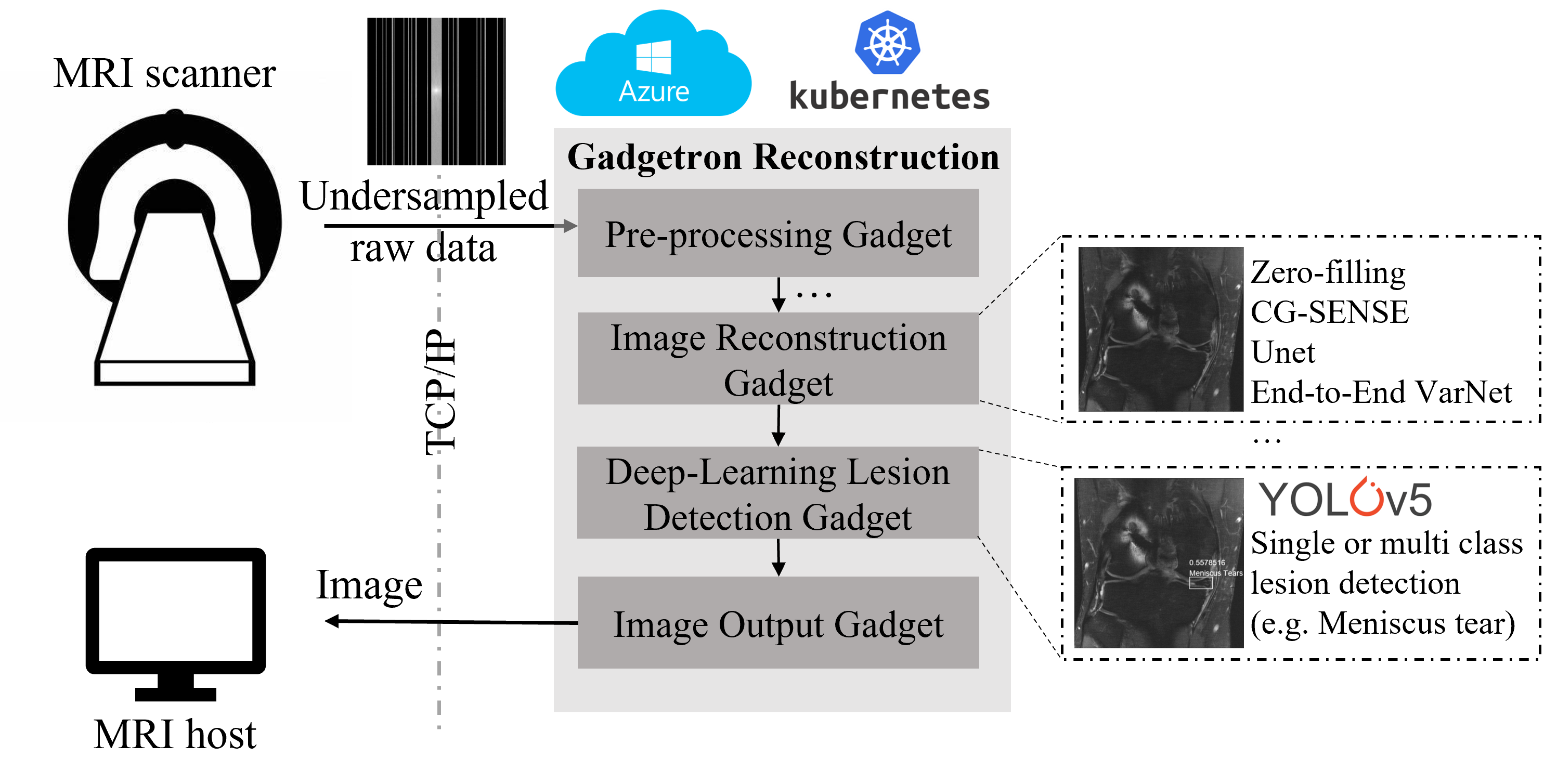}
\caption{Flowchart of the end-to-end image reconstruction and lesion detection pipeline. Different reconstruction methods can be applied in the image reconstruction gadget. The lesion detection gadget can also be extended with different single- or multi-class lesion detection algorithms, or combined with other classification methods.}
\label{fig:flowchart}
\end{figure}

\begin{figure*}
\includegraphics[width=0.6\textwidth]{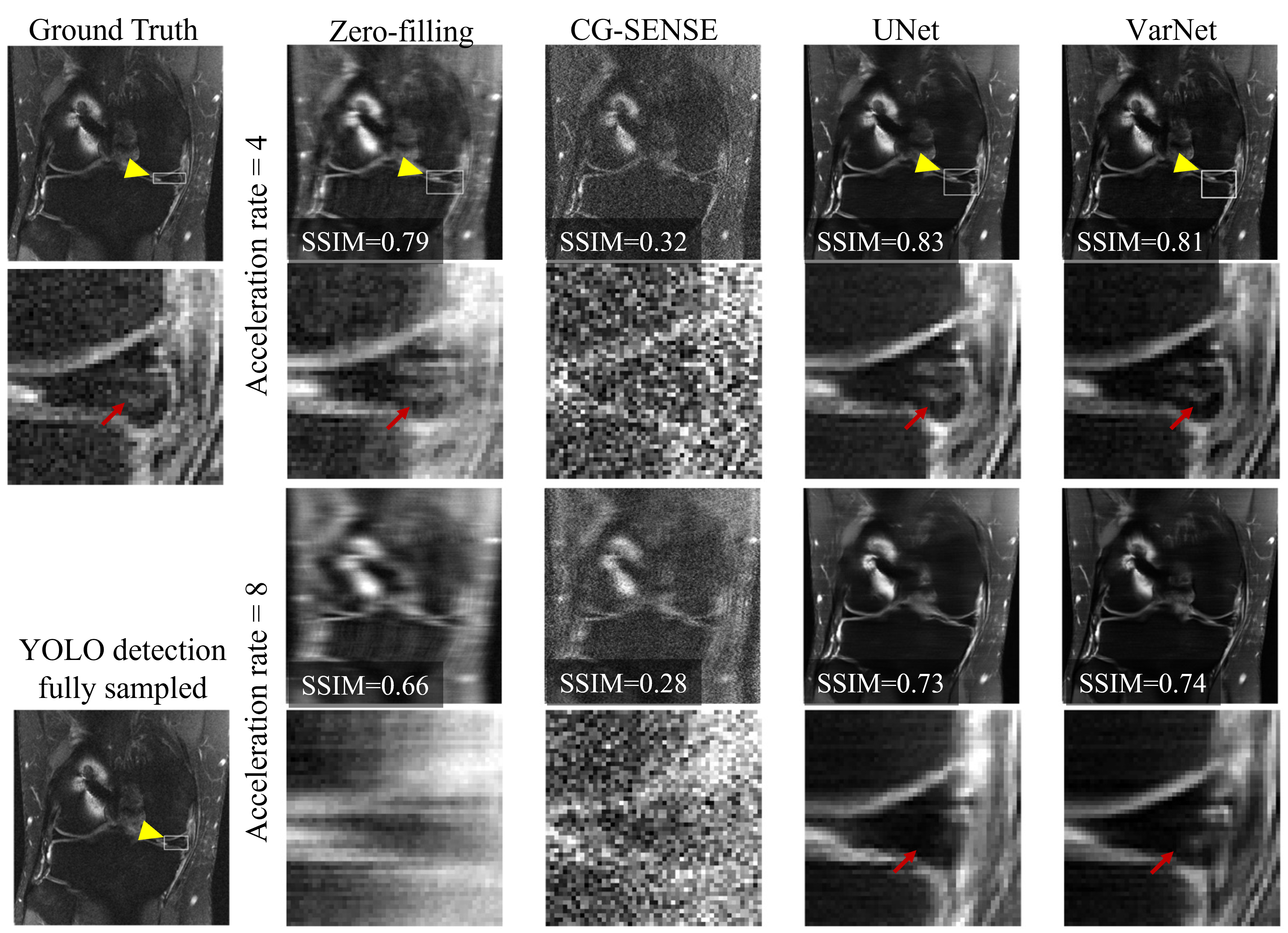}
\centering
\caption{An example slice with meniscus tear is shown with ground-truth bounding box and YOLO detected bounding boxes laid over the images (yellow arrowhead). Zoomed-in areas are shown with red arrow pointed to the meniscus tear. The detection network fails in CG-SENSE reconstruction due to its low SNR and in Zero-filling FFT at rate = 8 because of the artifacts. Comparing UNet and VarNet reconstruction at rate = 4 and rate = 8, both have detailed structures lost in the meniscus area at rate = 8, thus result in missed detection}
\label{fig:example}
\end{figure*}

The release of large  MRI raw datasets such as the fastMRI dataset \cite{zbontar2018fastmri} has advanced community engagement in tackling deep learning based MRI reconstruction challenges. However, research leveraging this open data resource has revealed drawbacks in many current DL methods \cite{knoll2020advancing, muckley2021results} owing, in part, to the use of  global quantitative metrics such as SSIM or NMSE for evaluation, which tends to optimize for good overall image quality (reduction of visible artifacts and improved signal-to-noise ratio). A chief challenge remains in that ability to reconstruct clinically important pathology is not considered and the fastMRI challenges show that deep learning techniques often fail to reconstruct fine details such as meniscus tear despite having good global quantitative metrics \cite{knoll2020advancing}. Hence, there is a need for clinically-aware deep learning techniques and workflows, which not only reconstruct images, but also detect lesions such as meniscus tears. Designing clinically-aware deep learning approaches requires datasets with the annotations of lesions for performance evaluation. Recently, the fastMRI+ dataset \cite{zhao2021fastmri}, which contains bounding box annotations for fastMRI dataset, was released for enabling this type of work. In this paper, we demonstrate an end-to-end pipeline, which includes image reconstruction and lesion detection.
 
In order to build a clinically-aware deep learning image reconstruction pipeline, it needs to provide clinically accepted latency. Current offline workflows need cumbersome data transfer processes and powerful local computing resources to handle the large raw datasets. But these extra steps limit large scale testing of image reconstruction pipelines and ultimately limit the clinical deployment of the developed methods. The Gadgetron \cite{hansen2013gadgetron} is an open-source framework capable of integrating DL reconstruction and lesion detection components in a way that can be leveraged for testing of multiple algorithms and facilitate seamless scanner integration.\\
{\indent}The proposed framework takes the undersampled raw MRI measurements in a form that could be produced directly from an MRI scanner as input and performs MRI reconstruction followed by lesion detection. The end-to-end framework is developed using Gadgetron framework and deployed in managed cloud computing services. Here we have compared the performance of conventional and deep learning based image reconstruction algorithms for the meniscus tear detection. To the best of our knowledge, this is the first study developing an automated MRI reconstruction and detection framework for raw-data to clinical pathology MRI reconstruction workflow leveraging a publicly available dataset.
\section{Methods}
\subsection{Background on MRI Reconstruction}
In MRI, data acquisition is performed in the spatial frequency domain, known as k-space. The forward model for the acquired k-space measurements are given as
\begin{equation}\label{Eq:Forward_Model}
{\bf f}_{\Omega} = {\bf A}_{\Omega} \mathbf{x} +{\bf n}, 
\end{equation}

where ${\bf f}_{\Omega}\in{\mathbb C}^{M}$  denotes the acquired subsampled k-space with $\Omega$ denoting the subsampling pattern, ${\bf A}:{\mathbb C}^{N} \to {\mathbb C}^M$ is the encoding operator containing partial fourier matrix and coil sensitivities, and ${\bf n}\in {\mathbb C}^{M}$ is the measurement noise. The image recovery from the acquired subsampled measurements is given as 
 \begin{equation}\label{Eq:inverse_problem}
\arg \min_{\bf x} \|\mathbf{f}_{\Omega}-\mathbf{A}_{\Omega}\mathbf{x}\|^2_2 + \cal{R}(\mathbf{x}),
\end{equation}
where first term denotes the data consistency and second term, $\cal{R}(\mathbf{\cdot})$, is a regularizer. The Eq.\ref{Eq:inverse_problem} has been conventionally solved by iterative optimization algorithms \cite{fessler_SPM}. Recently, data-driven and physics-guided deep learning techniques have emerged as an alternative technique for solving the inverse problem \cite{WangDLMRI,Hammernik,LeslieYing_SPM,Knoll_SPM}.

\begin{figure*}
\includegraphics[width=0.9\textwidth]{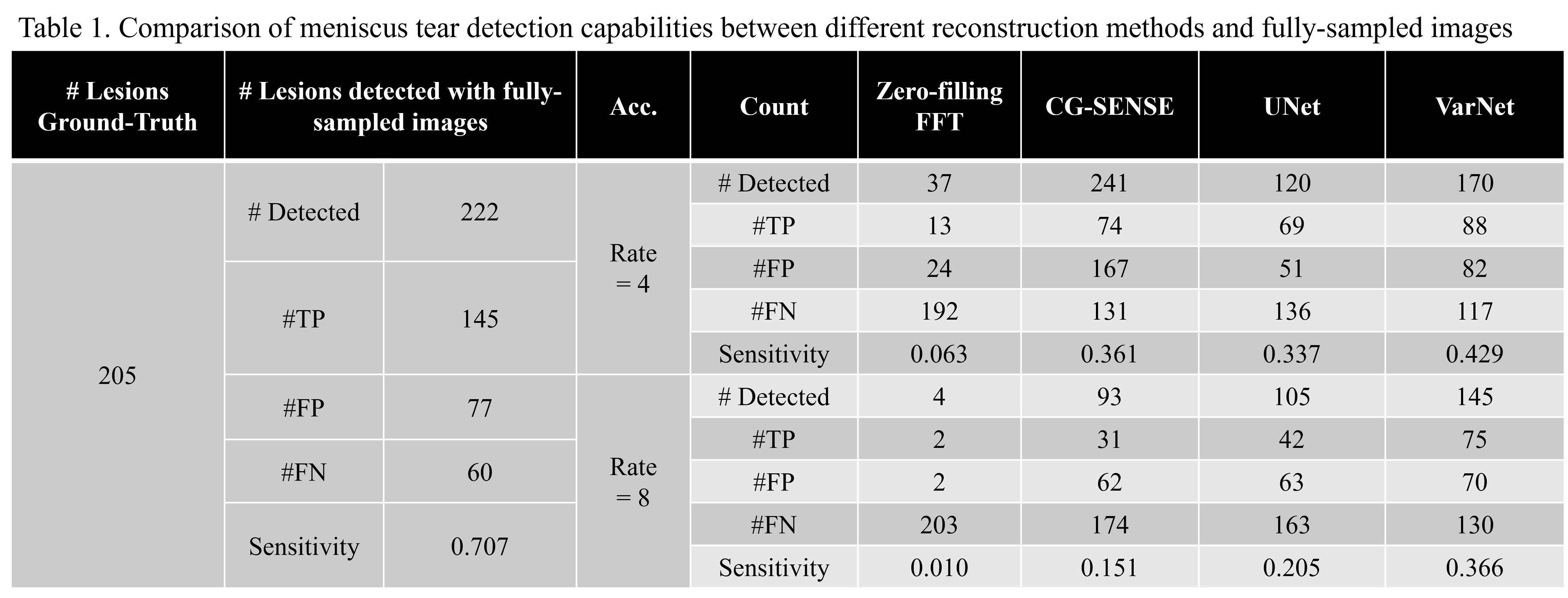}
\centering
\label{fig:metrics}
\end{figure*}
\subsection{Meniscus detection network with YOLO}
YOLO is a state-of-art, real-time object detection problem that integrates classifier and localizer into one neural network \cite{glenn_jocher_2021_4679653}. In our study, a detection network of meniscus tears based on the YOLO architecture was trained on fully-sampled knee MR images obtained from the fastMRI public dataset and the fastMRI+ clinical annotations \cite{zbontar2018fastmri, zhao2021fastmri}. The fully-sampled multi-coil raw data was reconstructed with an inverse Fast Fourier Transform (FFT) followed by root-sum-of-squares (RSS) coil combination. The images were stored in Portable Network Graphics (PNG) format for the training. A total of 7189 knee images were used in the training with 75\% having meniscus tears labeled and the other 25\% randomly selected from all the negative images (i.e., no meniscus tears) from the multi-coil knee dataset to serve as background images. The validation data had 870 images with the same background ratio and the test data had 1343 images from 37 subjects picked from the original fastMRI validation set. The detection model was trained with YOLOv5s architecture on Azure ML platform for 300 epochs using an Adam optimizer, and fine-tuned using the HyperDrive package \cite{glenn_jocher_2021_4679653}.\vspace{-5mm}

\subsection{Implementation with Gadgetron framework}
Gadgetron \cite{hansen2013gadgetron} is an open source framework for medical image reconstruction. It uses streaming data processing pipeline, which can receive TCP/IP connections from command line clients or MRI systems. Each Gadgetron pipeline consists a series of modules or "Gadgets", for instance, Fourier transform, which combine to form a complete medical image reconstruction pipeline. The Gadgetron has also been extended to support distributed computing which can provide a substantial reconstruction speed up \cite{xue2015distributed}. In this work, two Python Gadgets (image reconstruction Gadget and lesion detection Gadget) were implemented in order to run reconstruction with undersampled raw data and to detect lesion with reconstructed image, respectively. As shown in Fig.1, these two Gadgets were sequentially embedded in the Gadgetron framework. Both image reconstruction and lesion detection gadgets were parameterized such that different image reconstruction and lesion detection methods could be tested. The newly developed Gadgetron pipeline was containerized using Docker container and deployed in Azure Kubernetes Service (AKS) using open source deployment tools (https://github.com/Microsoft/gadgetron-azure).\vspace{-3mm}

\subsection{Imaging Experiments and Evaluation}
To evaluate the lesion detection capabilities of different reconstruction methods with the proposed pipeline, the pre-trained meniscus tear detection network was used in the lesion detection Gadget, and four different reconstruction methods (i.e., Zero-filling FFT, CG-SENSE, UNet, Model-based VarNet) were used in the image reconstruction Gadget, respectively \cite{Sense, maier2021cg,ronneberger2015u,sriram2020end}. The 37 test subjects with fully-sampled raw data were undersampled with randomly distributed mask at accelerate rate = 4 and 8, respectively. Then data of each subject was passed through the newly developed Gadgetron pipeline sequentially, each run four times with different reconstruction methods.

Test of the meniscus tear detection network was also run on fully-sampled images as reference. With the predicted bounding box information, the total number of detected meniscus tears, as well as the number of true positives (TP), false negatives (FN), and false positives (FP) relative to the ground-truth clinical labels and detection sensitivity were calculated.

We also calculated and compared the difference of SSIM metric between FP slices and FN slices for each reconstruction method, in order to assess the ability of SSIM to reflect lesion detection capability.
  
Finally, we trained the same meniscus tear detection network on images produced by the VarNet method, and ran test with both VarNet inference images and fully-sampled images to further assess the difference in reconstructed fine details.

\section{Results}
\label{sec:Results}
Images and zoomed-in areas of an example slice are shown in Fig.\ref{fig:example} with YOLO detected lesions marked with bounding boxes to compare the reconstruction quality and lesion detection capability of different reconstruction methods. VarNet, as the best performing reconstruction among the four, is able to preserve meniscus tear at rate = 4, but not at rate = 8.

\begin{figure}
\includegraphics[width=0.45\textwidth]{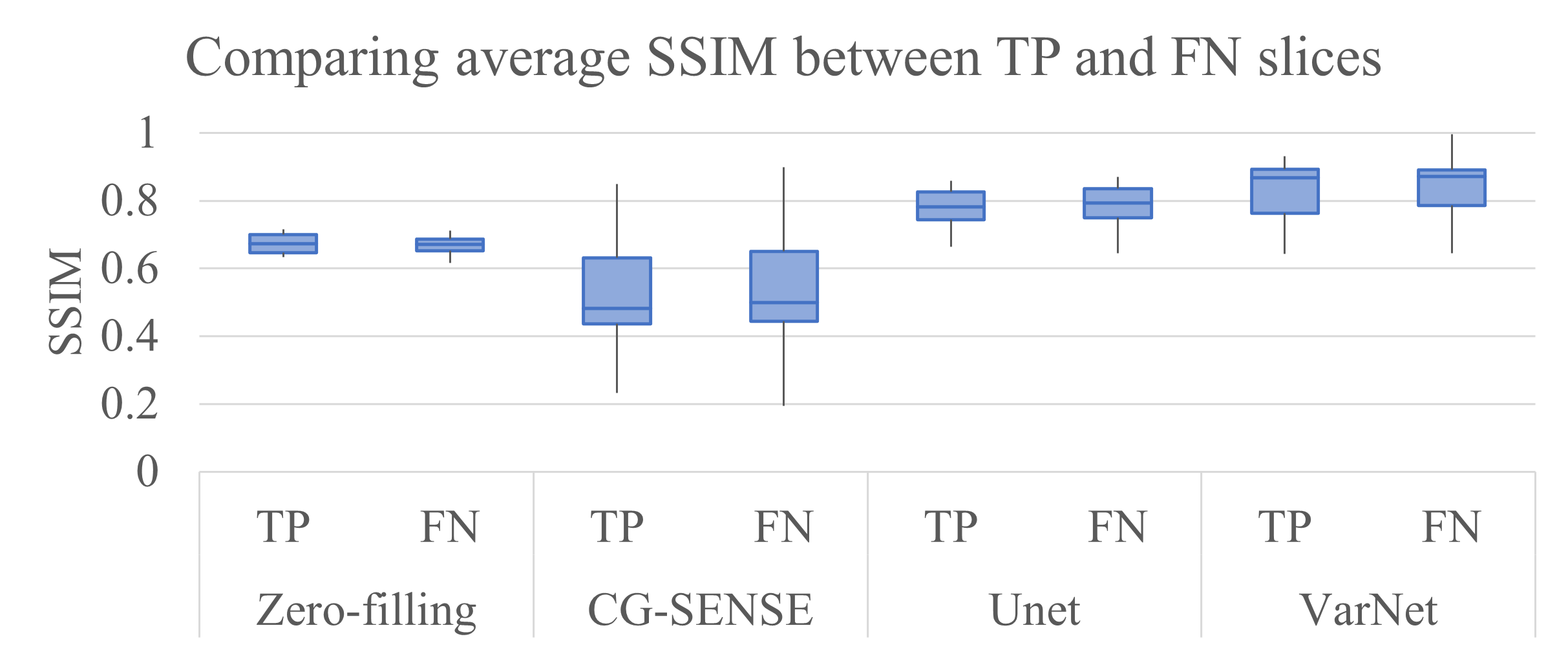}
\centering
\caption{Comparison of average SSIM between TP and FN slices with different reconstruction methods at acceleration rate = 4. No significant differences in SSIM are identified between TP and FN slices. R = 8 is not shown due to the limited number of TP slices}
\label{fig:ssim}
\end{figure}

Table 1 provides all the evaluated metrics for different reconstructions, compared to the YOLO detection performance on fully-sampled images. Fig.\ref{fig:ssim} shows comparison between average SSIM values and amount of TP\&FN for each  reconstruction method. We observe that the high SSIM values which are used as a performance criteria in the literature  do not differ for slices with TP\&FN. 

Table 2 presents the detected lesions for fully-sampled and VarNet inference test images at R=4 using the YOLO network trained with VarNet (R=4) reconstructed training images. In this scenario, number of lesions detected with Fully-sampled images exceed the reported values in Table 1. In other words, detection network outputs more false positives when it is trained with the VarNet images. 

\begin{figure}
\includegraphics[width=0.48\textwidth]{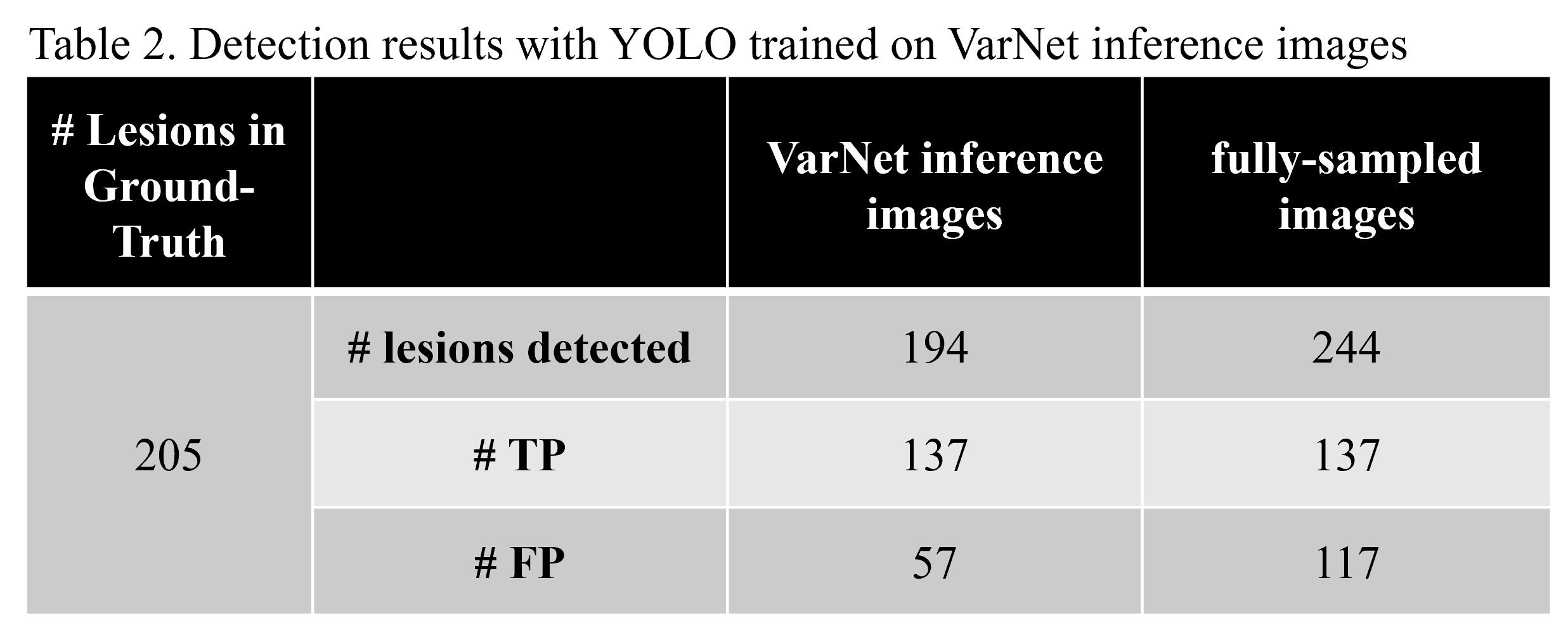}
\centering
\label{fig:varnetyolo}
\end{figure}
\section{Discussion and Conclusion}
In this study, we proposed an end-to-end deep learning framework, which enables MRI reconstruction and lesion detection in a single pipeline. The proposed framework enables rapid testing and comparison of image reconstruction algorithms not only in terms of global quality metrics but also in terms of their ability to preserve clinically important details. The framework was developed using Gadgetron, which facilitate direct connection to MRI system for clinical testing. 

Although deep learning approaches achieve visually improved reconstruction quality at high acceleration rates, fine details such as meniscus tears are either partially or completely lost during the reconstruction. As expected, YOLO detection algorithm did not detect the meniscus tears in such reconstructions. The proposed approach is a first step to automatic evaluation of deep learning image reconstruction algorithms. The proposed use of the YOLO network for lesion detection is not intended to represent the best or final lesion detection for this type of task, but it creates an automatic benchmark that allow us to compare two or more reconstruction algorithms in terms of their effect on a detection task. A human image reader might be affected differently by the changes in image quality and therefore the the results should require more careful study to assess clinical impact. However, we can conclude that highly accelerated deep learning reconstruction methods fail to reconstruct details that are critical for automatic detection meniscus tears even when the algorithms achieve good global quality metrics as measured by SSIM \cite{knoll2020advancing}. Hence, developing new clinically relevant metrics that can capture and provide important pathology information is warranted \cite{muckley2021results}. 
Another observation found that training YOLO with deep learning based accelerated MRI images rather than fully-sampled images might lead to overfitting as the detection network could map to lesions that may have already been lost during the reconstruction. Hence, developing texture and detail preserving reconstruction architectures can be beneficial for more precise lesion detection.\\
{\indent}In this study, the YOLO lesion detection network was only trained on meniscus tears in order to focus on evaluating the reconstruction quality on fine details. However, our model is noticed to mistakenly recognize other types of lesions (e.g. displacement of meniscal tissue) as meniscus tears due to their similarity. Thus, the current lesion detection model can be further extended to a multi-class detection network for more comprehensive evaluation of different diseases given in the fastMRI+ dataset \cite{zhao2021fastmri}. The fastMRI+ annotations also have important limitations as detailed in the description of the annotations. Specifically, the annotations were produced by a single reader and careful review of the annotations maybe be needed especially when considering other lesions that are less abundant in the dataset.\\
{\indent}In conclusion the presented work focused on deomonstratinig the feasibility of automating MRI process by using existent state-of-the-art reconstruction and detection processes and deploying them all-together in Gadgetron. We believe observations and challenges presented in our study can inspire for future work on developing more rigorous metrics that capture an image reconstruction methods ability to preserve the fine detail needed for lesion detection and clinical evaluation.\vspace{-5mm}

\section{Compliance with ethical standards}
\label{sec:ethics}
This research study was conducted retrospectively using human subject data made available in open access \cite{zbontar2018fastmri, zhao2021fastmri}. Ethical approval was not required as confirmed by the license attached with the open access data.

\bibliographystyle{IEEEbib}
\bibliography{strings,refs}

\end{document}